\documentclass{article}

\usepackage[final]{neurips_2021}

\usepackage{amsmath}
\usepackage{graphicx}

\usepackage{pdflscape}
\usepackage{subfigure}
\usepackage{caption}
\usepackage{todonotes}
\usepackage[utf8]{inputenc} 
\usepackage[T1]{fontenc}    
\usepackage{hyperref}       
\usepackage{url}            
\usepackage{booktabs}       
\usepackage{amsfonts}       
\usepackage{nicefrac}       
\usepackage{microtype}      
\usepackage{xcolor}         

\newcommand\norm[1]{\left\lVert#1\right\rVert}
\usepackage{natbib}
\title{Deep Reinforcement Learning Policies\\for Underactuated Satellite Attitude Control}

\author{%
  Matteo~El Hariry\\
  Argotec srl\\
  \texttt{matteo.elhariry@argotecgroup.com} \\
\And
  Andrea~Cini\thanks{Now at IDSIA, Universit\`a della Svizzera italiana.}\\
  Argotec srl\\
  \texttt{andrea.cini@argotecgroup.com} \\
   \And
  Giacomo~Mellone\\
  Argotec srl\\
  \texttt{giacomo.mellone@argotecgroup.com} \\ 
 \And
 Alessandro~Balossino\\
  Argotec srl\\
  \texttt{alessandro.balossino@argotecgroup.com} \\
}

\begin{document}

\maketitle

\begin{abstract}
  Autonomy is a key challenge for future space exploration endeavours. Deep Reinforcement Learning holds the promises for developing agents able to learn complex behaviours simply by interacting with their environment. This paper investigates the use of Reinforcement Learning for the satellite attitude control problem, namely the angular reorientation of a spacecraft with respect to an inertial frame of reference. In the proposed approach, a set of control policies are implemented as neural networks trained with a custom version of the Proximal Policy Optimization algorithm to maneuver a small satellite from a random starting angle to a given pointing target. In particular, we address the problem for two working conditions: the nominal case, in which all the actuators (a set of 3 reaction wheels) are working properly, and the underactuated case, where an actuator failure is simulated randomly along with one of the axes. We show that the agents learn to effectively perform large-angle slew maneuvers with fast convergence and industry-standard pointing accuracy. Furthermore, we test the proposed method on representative hardware, showing that by taking adequate measures controllers trained in simulation can perform well in real systems. \end{abstract}

\section{Introduction}

In the past few years, small satellites commonly designed to operate at Low Earth Orbits (LEO) are increasingly being deployed in Deep Space for exploration and scientific observation, especially as a secondary payload in bigger missions. The wide variety of possible scenarios and environments that these systems interact with during their operational life inevitably leads to faults/failures~\citep{henna2020fault}. Failures can be categorized according to their pertinence level, namely the system, the sensors or the actuators. In this paper, we focus on the latter category: we investigate the use of reinforcement learning to cope with the failure of reaction wheels (RWs). In case of RW failure, the system, having a lower number of actuators than degrees of freedom, becomes \textit{underactuated}. Underactuated control is challenging for classic control strategies since it imposes constraints on the trajectories that are possible to follow in the configuration space. Since hardware redundancy of reaction wheels is not always a possible solution for satellites, increasing the robustness of the control algorithms arguably represents the most cost-effective solution to enhance the overall system reliability. In this setting, standard advanced techniques often rely on a model to plan and perform trajectory optimization. Reinforcement Learning (RL) \citep{sutton2018reinforcement}, on the other hand, is \textit{model-free}, i.e. it can learn a control policy, without an explicit model of the system, through direct interaction with the environment. Deep Learning~\citep{lecun2015dl}, in particular, has brought to new exciting results in the application of RL to control problems~\citep{heess2017elbre,openai2018ldm, rubik-cube21}.

In this work we use state of the art RL for~(underactuated) spacecraft attitude control. Control actions are continuous torques applied to the principal body frame axes. Our novel contributions are the following: first, the implementation of $10$ successful RL based controllers, $1$ nominal agent, and $9$ agents in charge of aligning the satellite with a specific direction in space, despite the failure of an actuator. For the underactuated case we devised 3 agents, each specialized in aligning the spacecraft with one axis, for every possible failure.  Second, the validation of the aforementioned controllers on the \emph{Argomoon}~\citep{ditana2018agm} small satellite platform with hardware-in-the-loop tests. The integrated systems will undergo in-orbit validation during the secondary phase of the Argomoon deep space mission after deployment from NASA's \textit{Space Launch System}. The launch is due in 2022. Here, we show how the models trained in simulation can transfer their learned policies performing well on the target HW.

In the following, we discuss related works in Section~\ref{sec1} and provide details on satellite attitude control in Section~\ref{sec2}. In Section~\ref{sec3}, we describe the methodology, modelling the problem as Markov Decision Process~(MDP). In Section~\ref{sec4} we present results for the attitude control problem both in simulation, and in a relevant real-time system based on ground-model components of a real spacecraft. Finally, we draw our conclusions in Section~\ref{sec5}, discussing possible future works.


\section{Related work}\label{sec1}

\paragraph{Underactuated control} The problem of spacecraft attitude control with two reaction wheels is a particularly challenging problem, subject of active research.
Despite progress in the angular velocity stabilization problem for the case of actuator failures~\citep{tsiotras2000csaf}, complete attitude control is still an open problem~\citep{heess2017elbre} and, even if both time varying and time-invariant feedback controllers~\citep{gui2013amc,chaurais2015acus} have been devised, they work only under specific assumptions with limited performances, i.e. not covering complete angular maneuvers. Even though it is still not clear how to solve this problem, interesting approaches exist. \citet{zavoli2017sapus} present a smooth time-invariant wheel rate command under ideal conditions of zero total angular momentum and no reaction wheel saturation, proving asymptotic convergence to a restricted region around the target. \citet{tsiotras2000csaf} provide interesting results for the underactuated attitude stabilization problem for a non-symmetric spacecraft with gas jet actuators. \citet{tsiotras1995acass} derive a feedback control law able to stabilize the spacecraft to a circular attractor~(corresponding to a steady rotation about the axis with no control authority).

\paragraph{Reinforcement Learning for Attitude Control} Attitude control methods based on RL have been proposed with different objectives. \citet{elkinsa2020autonomous} managed to reach very high pointing accuracy training a PPO agent with discrete action space. \citet{vedant2019rl_ac} trained RL agents to control a family of satellites with mass ranging from $0.1$ to $100000$ kg. \citet{xu_model-based_2019} use model-based DRL with heuristic search to obtain an adaptive attitude control strategy. 

To the best of our knowledge, none of previous works tackled underactuated satellite control with RL; furthermore none of them actually implemented the RL controller in real space-graded hardware.

\section{Background}\label{sec2}

\subsection{Satellite Attitude Dynamics}

The attitude of a rigid body describes its angular orientation with respect to a reference frame and it is often formalized either by Euler angles, which define the three elemental rotations along the three body-frame axes, namely pitch, roll, and yaw, or by quaternions, which allow to represent the relative orientation of two coordinate systems as the angular rotation $\theta$ around a fixed vector $\overset{\rightarrow}{v}$. More details on the use of quaternions for attitude representations are provided in Appendix~\ref{appendix-quaternion}.

To simulate the attitude evolution of the satellite the Euler's rotation equations are used:
\begin{equation}
    \overset{\rightarrow}{M} = \textit{\textbf{I}}\dot{\overset{\rightarrow}{\omega}} + \omega{\times} \textit{\textbf{I}}\overset{\rightarrow}{\omega}\\
    \label{eq:euler}
\end{equation}
where $\textit{\textbf{I}}$ is the spacecraft's rotational inertia matrix with respect to the center of mass, $\omega$ and $\dot{\omega}$ are the angular rotation speed and acceleration vectors about the body-fixed principal axes, $\omega^{\times}$ is the skew-symmetric cross product matrix, while $M$is the external applied torque on the rigid body. The attitude is propagated using the Euler method for a fixed time $\Delta$t, which in our case corresponds to the system response time, i.e., the elapsed time between sending the command and receiving the telemetry (containing the new attitude information). More details are provided in~Appendix~\ref{appendix-propagation}.

\subsection{Satellite Attitude Control}

In general, the attitude control modes for a satellite can be different according to the mission phase. We do not consider particular tasks such as the attitude stabilization after the deployment from the carrier, nor the contingency or special modes required by specific mission events such as docking, presence of external disturbances or variations in the system's inertial mass. Instead, we assume the satellite already being in its final orbit and requiring the reorientation maneuvers to point towards a given target, for example to send/receive radio-signals, recharge batteries through solar panels, or acquire pictures. The system's momentum is particularly important for the detumbling operations, i.e. the maneuvers aimed at stopping the rotations of a spinning spacecraft, which are often dealt with by a dedicated controller. Since our focus is not detumbling, we assume zero speed on the reaction wheels and zero angular momentum as starting condition. We used these assumptions for both simulation and Hardware-in-the-Loop (HiL). 

Among different actuators that can be used to control the satellite's attitude, we focus on reaction wheels, which offer high pointing accuracy and work without relying on a fixed budget source, such as gas-thrusters, or on environmental factors such as the magnetorquers. We consider a system with three available RWs in the nominal case, then only two in the underactuated case, which can occur after failure of one wheel for a non-redundant control system hardware, or in case of multiple failures for redundant systems. The failure of the actuators is considered as a complete lack of response to the input signals, and is simulated by outputting constant zero torque along the predefined axis despite the agent's control signal.



\section{Problem formulation}\label{sec3}

Here we formalize the Attitude Control problem as a Markov Decision Process. 

\paragraph{State space.} The environment state space, and consequently the agent input, is constituted by the current attitude of the simulated spacecraft, i.e. the orientation quaternion $q_{0},q_{1},q_{2},q_{3}$ plus the angular rates along the three axes $\omega_{x},\omega_{y}, \omega_{z}$, and the RWs speeds $rw_{x}, rw_{y}, rw_{z}$.

\paragraph{Action space.} The action space is a $3$-dimensional vector representing the torques command for the RWs, $M_{x},M_{y},M_{z}$. 

The command provide a continuous torque for a $\delta t$ of $0.5$ s at each step for each of the three main body axes. To reduce energy consumption and the risk of early saturation of the reaction wheels, we limit the torques in the range $[-2, +2]~mN\cdot m$, which corresponds to $50\%$ of the maximum torque applicable.

\paragraph{Reward function.} The design of the reward function is fundamental to ensure that the desired behavior is obtained and that the payoff signal is such that the RL agents is able to learn. We design a dense reward function inversely proportional to the distance from the target, bounded between $[0, 0.5]$. In addition, we impose soft constraints on the final control policy by giving high negative reward to states outside of some predetermined nominal working conditions, e.g., a reward of $-1$ is associated to all the states where angular speed along any axis is over $0.1$ rad/s. Furthermore, we add a penalty term $(p)$ proportional to the sum of applied torques, to encourage learning energy efficient policies. Once the target orientation is reached, the agents receive a constant reward equal to $1$ for each step in which the pointing accuracy is within a predefined threshold. The set of constraints were selected after noticing how the initial policies trained with the proportional signal as the only reward failed to achieve smooth and stable control. Note that here we consider non-episodic settings: the agent must learn not only how to reach the target, but also how to stabilize its attitude and lock on the pointing target. To sum up, the reward function is:

\begin{equation}
    Reward =   \begin{cases}
                 -1 \hspace{3.5cm} if~any(\omega)~>~0.1~rad/s~\\
                 \\
                 +1 \hspace{3.5cm} if~\theta~<~thrs \\
                 \\ 
                \frac{1}{2} \left( 1 - \left( \frac{ \theta - thrs}{\pi}\right) ^{0.6} \right) - p \hspace{0.5cm} otherwise.
                \end{cases}
    \label{eq_reward}
\end{equation}

\paragraph{Design considerations.} The state definition does not include the current pointing target. In fact, we simply consider the quaternion $[1, 0, 0, 0]^{T}$ as target to reduce the input dimensionality: any pointing target can still be achieved by the final system by a change of reference. In practice, each simulation starts from a random pointing direction and the agent learns how to reach the reference one. Since this work is meant to provide a solution to be integrated directly in the control software of the satellite on-board computer, we constrain the control loop to a low frequency to not starve the other ongoing process in the satellite on-board computer. In particular, the agent perceives an observation and performs an action with a frequency of $2$ Hz. To provide control agents robustness against possible delays, we trained the models injecting in the control loop random delays sampled in the range [$0.5, 1$]s and added the last control torque command to the state vector.

The complete architecture of the proposed control scheme is composed by one nominal and $9$ underactuation controllers: the reference platform has $3$ RWs subject to failures and for each of the $3$ underactuated conditions we want a controller specialized in aligning one of satellite’s $3$ reference axes to a target direction.

\section{Methodology}

 We used a simulated environment to extensively evaluate the agents. We developed the environment by using the OpenAI gym interface~\citep{openAI_gym} and modelling the dynamics of the satellite while in orbit. As already mentioned, we use the ArgoMoon~\citep{ditana2018agm} platform as reference. Some of the physical characteristics of the satellite are provided in Table~\ref{tab:agm-sat}. At each environment reset, we sample a different starting orientation. Physical constraints are based on the reference platform and its physical limits. For example, we set the saturation speed of each reaction wheel is $7000$ rpm (hard constrain).

\paragraph{Agents training}  We generate the initial pointing direcation by sampling a random angle $\theta$ between [$30$, $180$]$^\circ$ and which is then converted into a quaternion by using spherical coordinates and the equivalence shown in the appendix. During the first episodes of learning we bias the sampling toward smaller angles during the first episodes. We train $10$ agents with different seed for every final controller, and we select the best ones according to the highest reward reached over the $40$ training epochs. For the underactuated case, the controller for the z-failed and y-aligned axis showed smallest average cumulative reward ($891.28$) and highest cumulative reward variance ($3.67$), while the x-failed and x-aligned controller reached the best average ($950.77$) and smallest reward variance ($17.88$). The difference is due to the satellite having a different inertia along the different axis. Each epoch consists of $15000$ steps in the environment and fix the length of each simulated episode to $500$ and $800$ steps for the nominal and underactuated case respectively. The simulation episode is either ended when the horizon is reached or when the angular velocity magnitude exceeds $0.1$ rad/s~(which in practice would imply the failure of the attitude determination system of the satellite). The controllers are implemented with Pytorch~\citep{pytorch2019} as Feed-forward Neural Networks, with $2$ hidden layers of $64$ neurons each, and trained on a AMD Ryzen Threadripper 1920X CPU. The policies were learned by using a custom policy gradient algorithm based on PPO~\citep{schulman2017ppo}. Relevant hyperparameters are in the appendix \ref{appendix-hyperparam}. With the above setup, training a batch of ten controllers takes approximately 26 hours.

\paragraph{Agents evaluation} Each controller has two main objectives: first, to minimize the time the satellite takes to align with the target direction. Second, to maintain an accurate pointing (w.r.t. the distance from the wanted attitude). The performance of the agents are evaluated based over $3000$ steps. In the underactuated case, the accuracy is considered w.r.t. the angular distance between the target vector and the main axis of the satellite to align. While, in the nominal case, we consider the angular distance between the target quaternion and satellite reference quaternion. Considering the platform specifications, besides the set of acceptable working conditions of the payloads, we set the threshold of the angular distance from the target to $0.01$ rad (i.e. $0.57$ degrees). Since it is extremely difficult to prove the convergence and stability properties of a controller based on neural networks, we statistically analyze the reliability and the performance of the proposed solution. Moreover, since the space industry is particularly risk adverse, we analysed both the mean and the "worst" performance of each controller at each step, considering the maximum angular distance from the target at each step across episodes. As shown in Figure~\ref{failed-plots} the mean performances of each agent are given by the blue line, while the blue area represents the standard deviation and the red area is given by plotting over each step the worst (most distant from the target) position among all the $10000$ evaluated episodes. This representation is particularly helpful in highlighting the overall behaviour of each final agent.

\begin{table}
\centering
\begin{tabular}{ll}
\hline
\textbf{Sat Inertia}& {$[0.19, 0.23, 0.17]~kg \cdot m^{2}$} \\
\hline
\textbf{RWs Inertia}& {$1.82 \cdot 10^{-05}~kg \cdot m^{2}$} \\
\hline
\textbf{Max RWs Torque}& {$0.004~N \cdot m$} \\
\hline
\textbf{Volume}& {$6U (30 \cdot 20 \cdot 10~cm^{3})$} \\
\hline
\end{tabular}
\caption{Reference satellite parameters}
\label{tab:agm-sat}
\end{table}

\section{Final Results}\label{sec4}

The results shown are from two different evaluations: the first test is performed in the simulated environment using the trained control agents over $10000$ different episodes, while the second test evaluates the behavior of the same policies on the target hardware, where the trained neural networks and control scheme are used at test time after being ported to the C language and integrated in the satellite's real-time operating system.

\begin{figure}
\centering
\includegraphics[width=9cm, height=6cm]{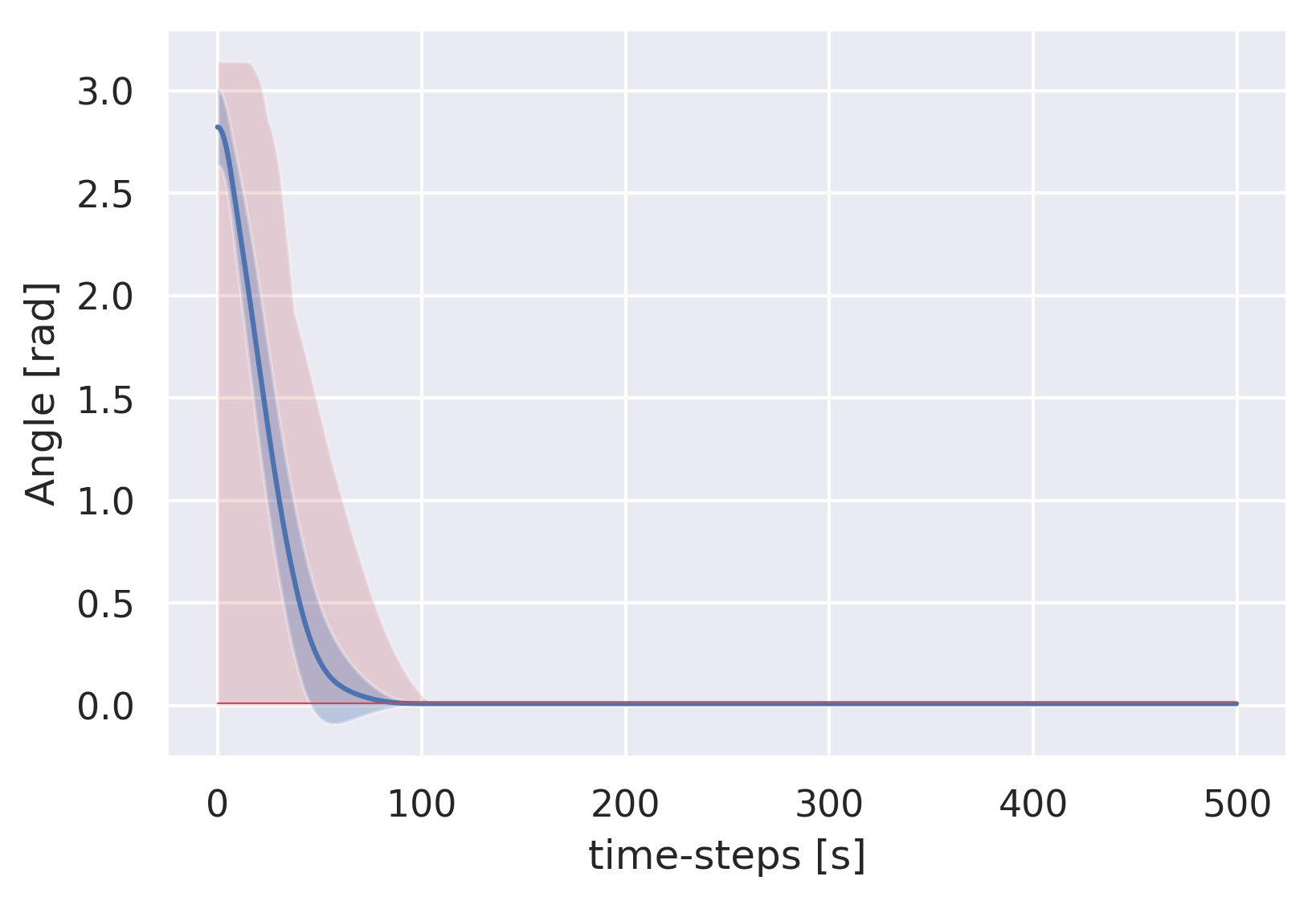}
\caption{Agents control history for the nominal case with average and variance performances in blue line and blue area, and absolute worst case performances highlighted by the red area. The statistics are obtained over 10000 runs and show how both the average and the worst case trajectory converge to the 0.01 rad threshold in less than 100 seconds.}
\label{nominal-plot}
\end{figure}

\begin{figure}
\centering
\subfigure[Failure on x-axis, align x]{\includegraphics[width=4.6cm, height=4.5cm]{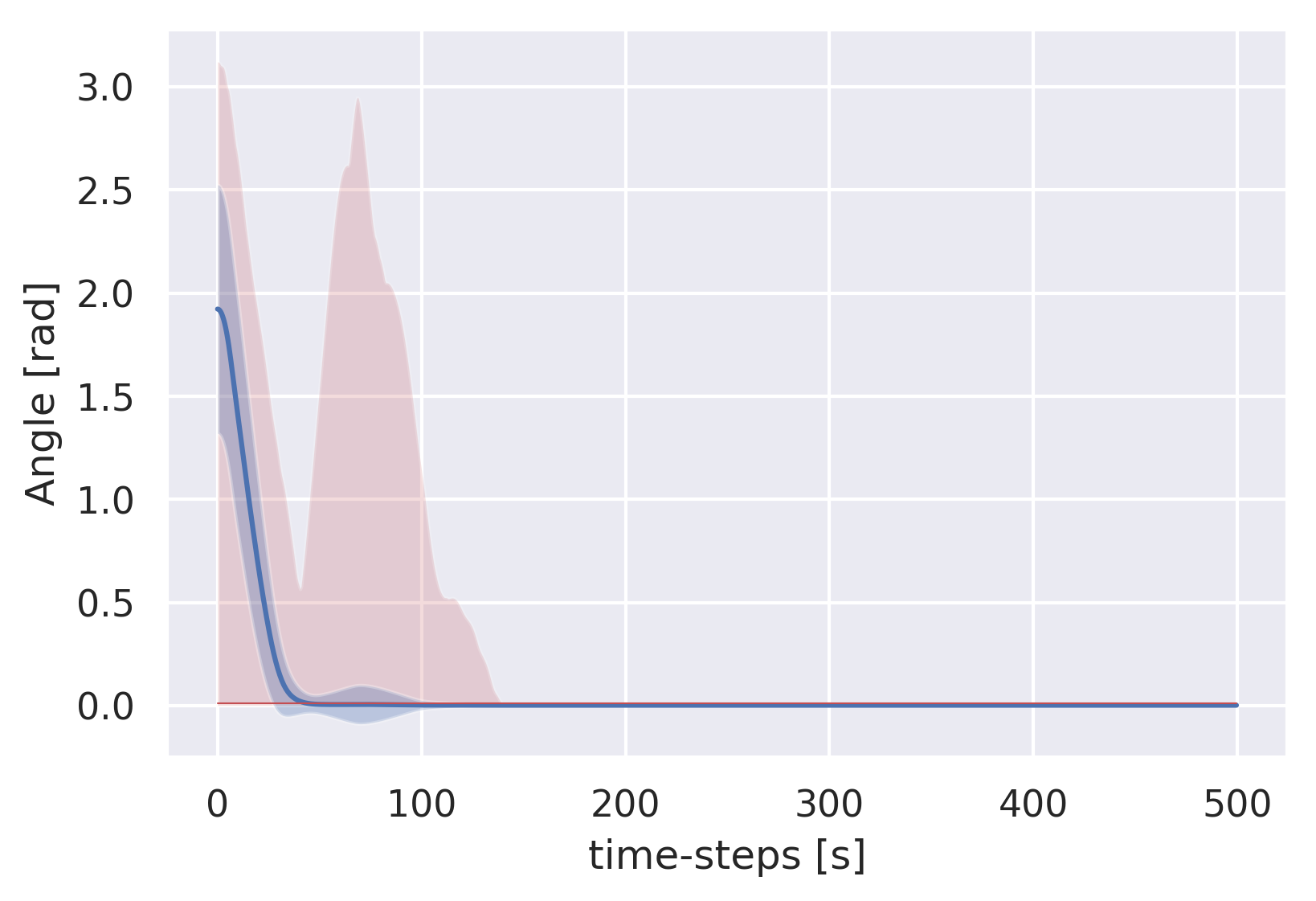}}
\subfigure[Failure on x-axis, align y]{\includegraphics[width=4.6cm, height=4.5cm]{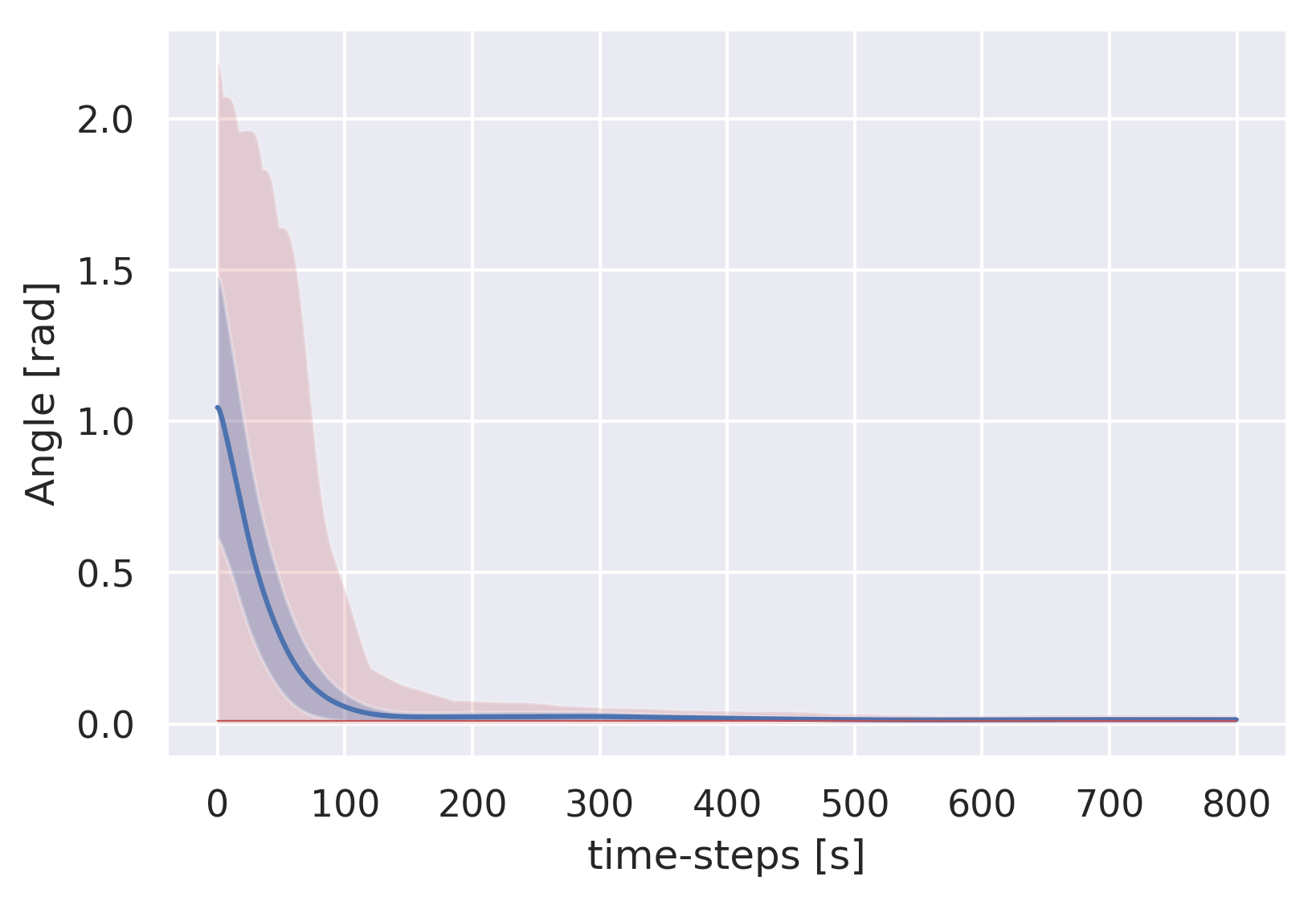}}
\subfigure[Failure on x-axis, align z]{\includegraphics[width=4.6cm, height=4.5cm]{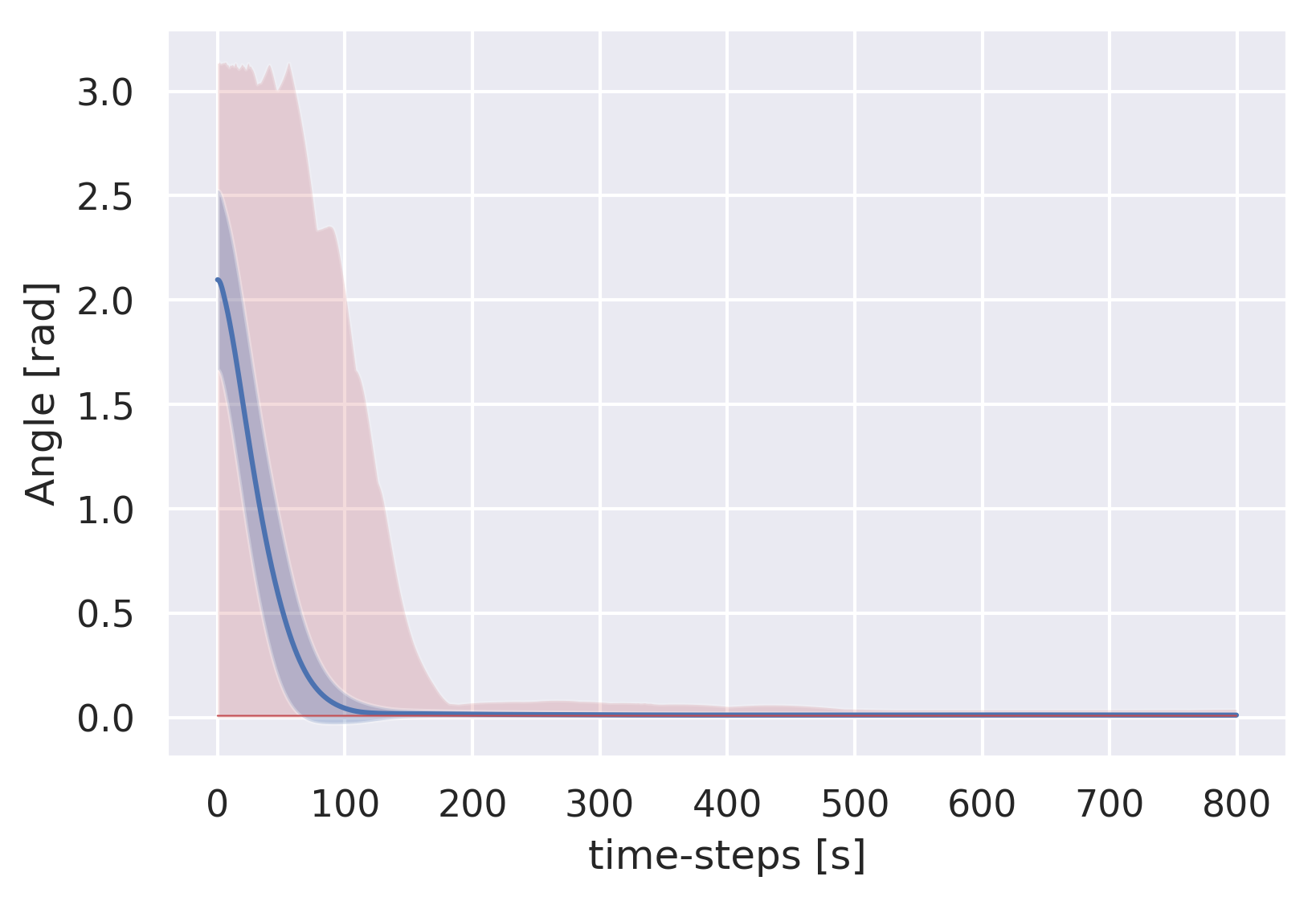}}
\vskip1ex
\subfigure[Failure on y-axis, align x]{\includegraphics[width=4.6cm, height=4.5cm]{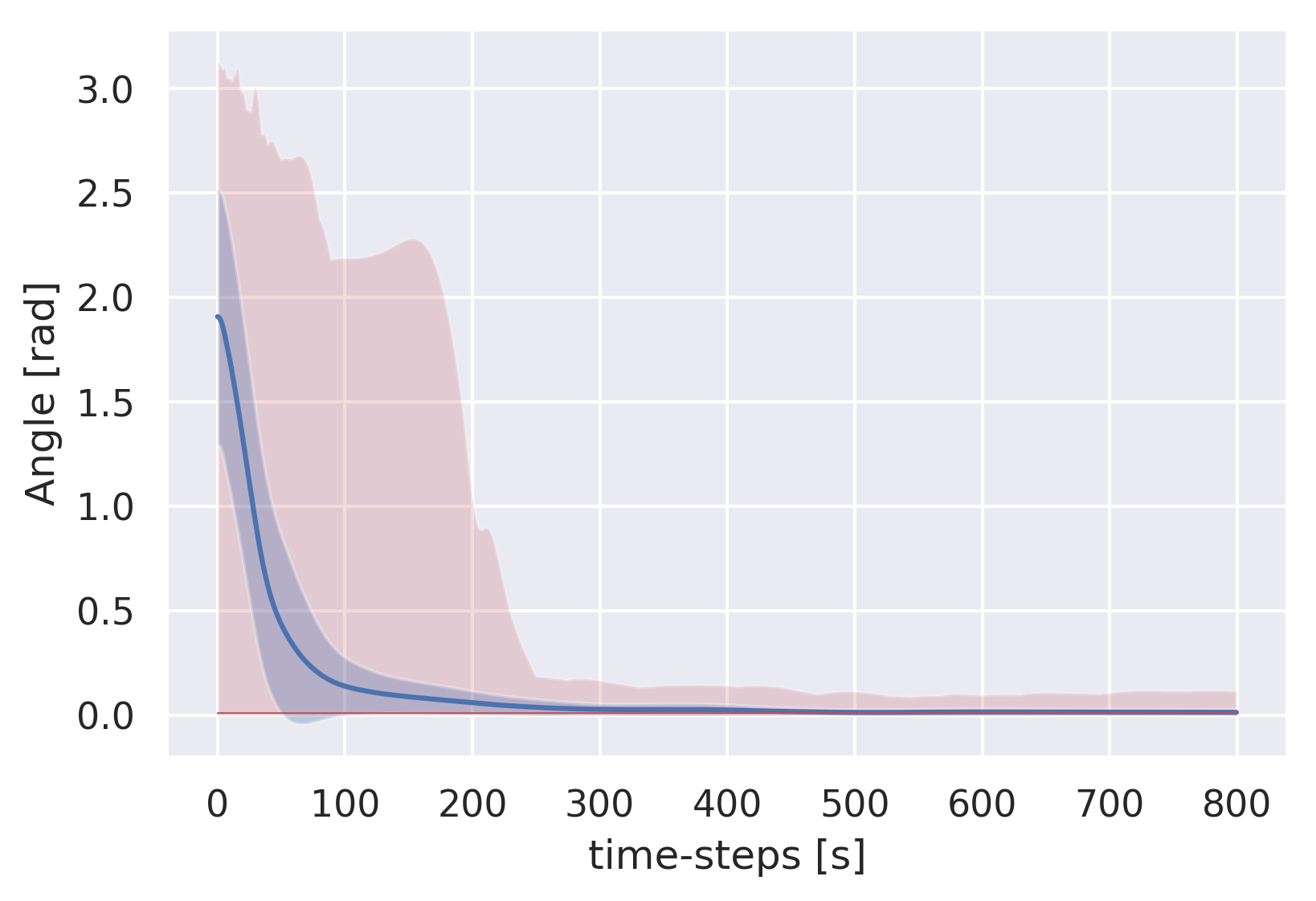}}
\subfigure[Failure on y-axis, align y]{\includegraphics[width=4.6cm, height=4.5cm]{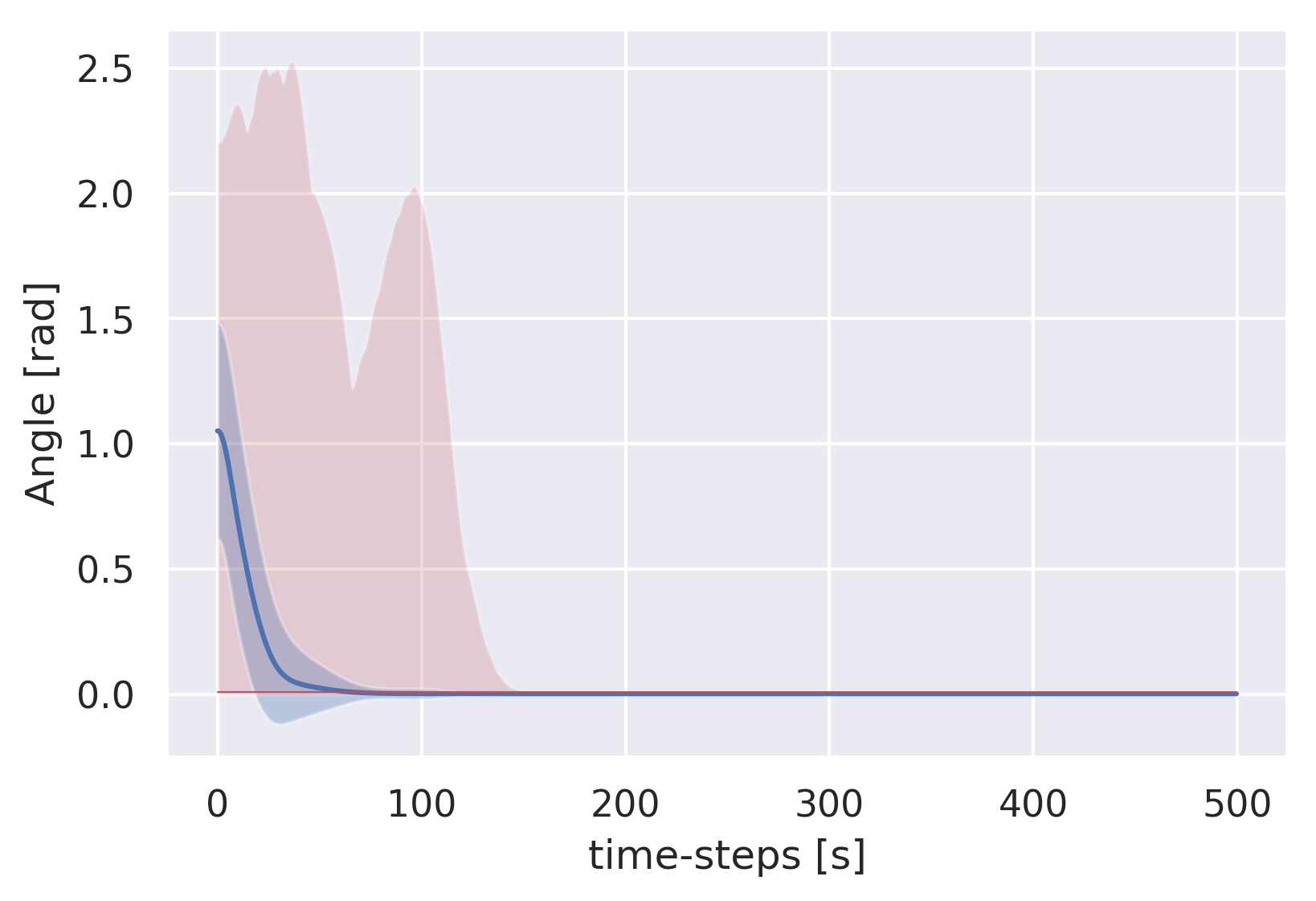}}
\subfigure[Failure on y-axis, align z]{\includegraphics[width=4.6cm, height=4.5cm]{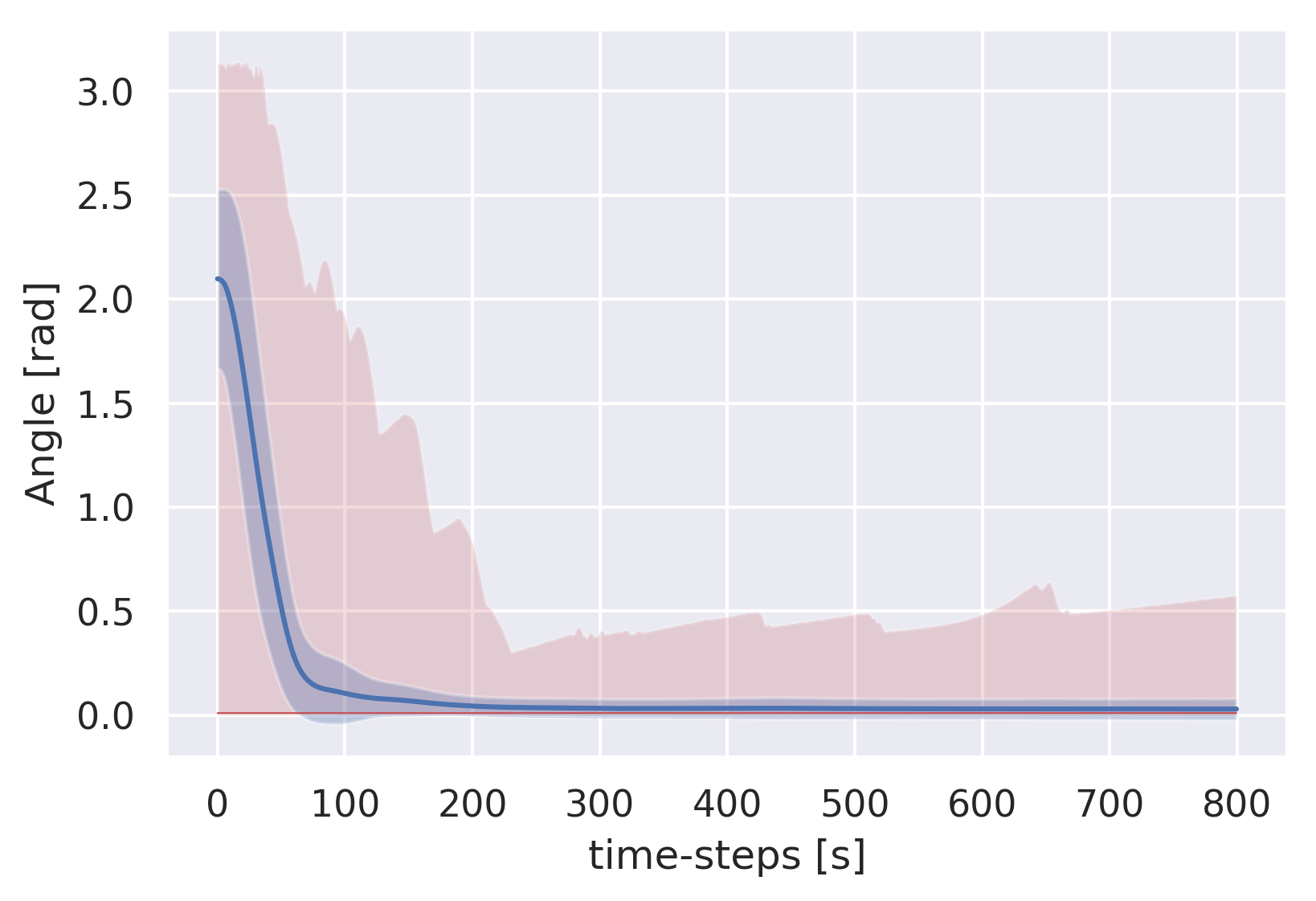}}
\vskip1ex
\subfigure[Failure on z-axis, align x]{\includegraphics[width=4.6cm, height=4.5cm]{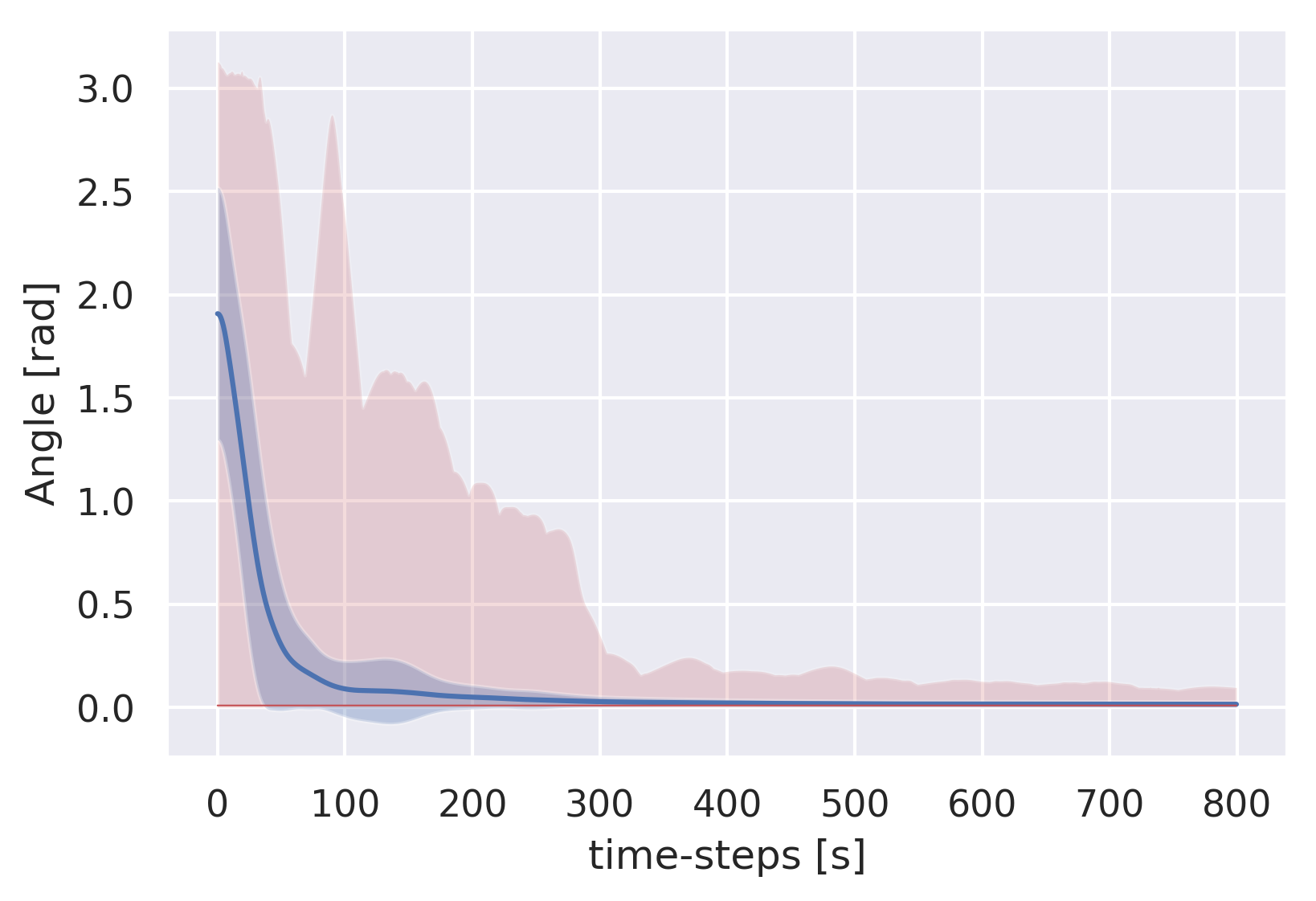}}
\subfigure[Failure on z-axis, align y]{\includegraphics[width=4.6cm, height=4.5cm]{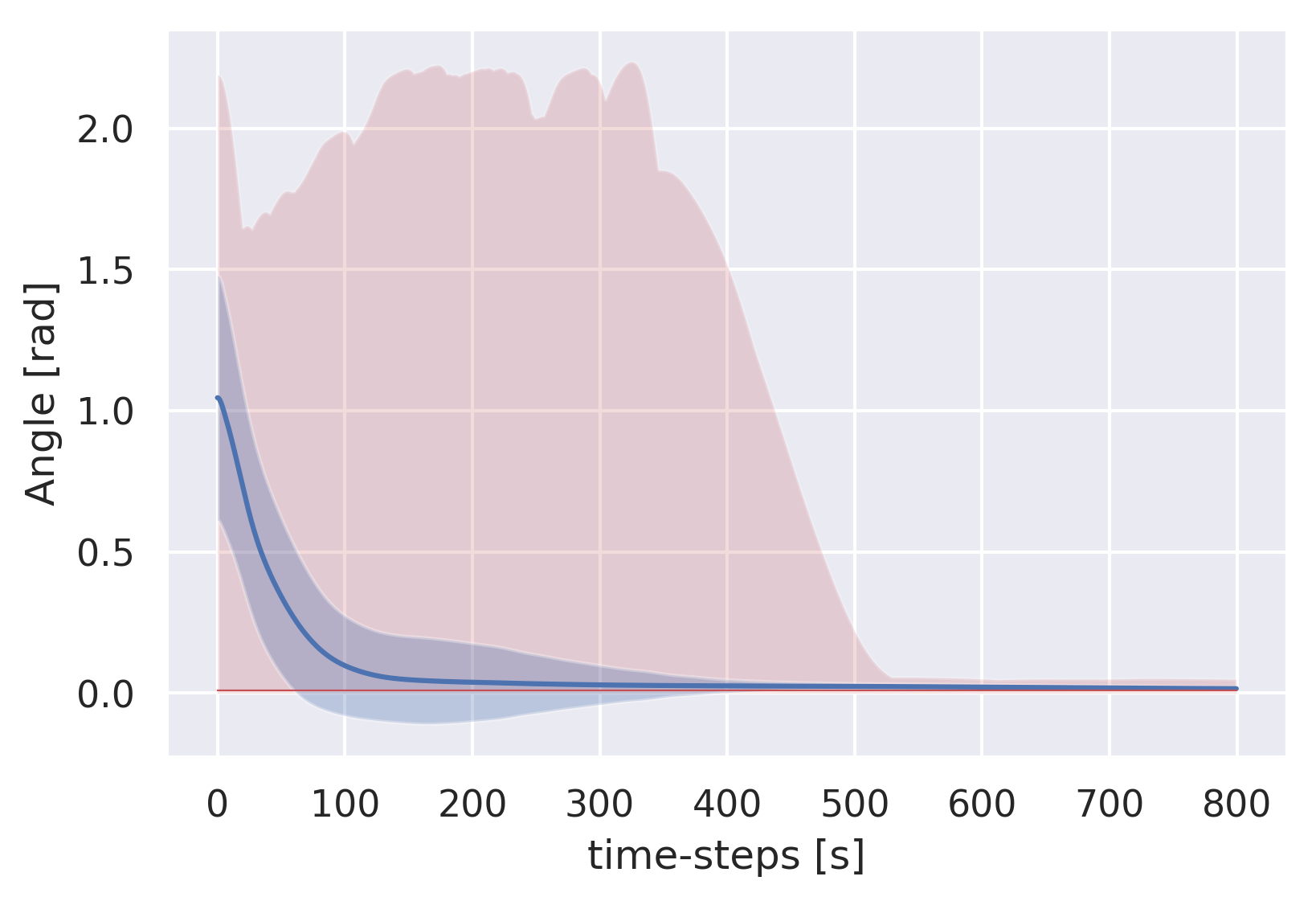}}
\subfigure[Failure on z-axis, align z]{\includegraphics[width=4.6cm, height=4.5cm]{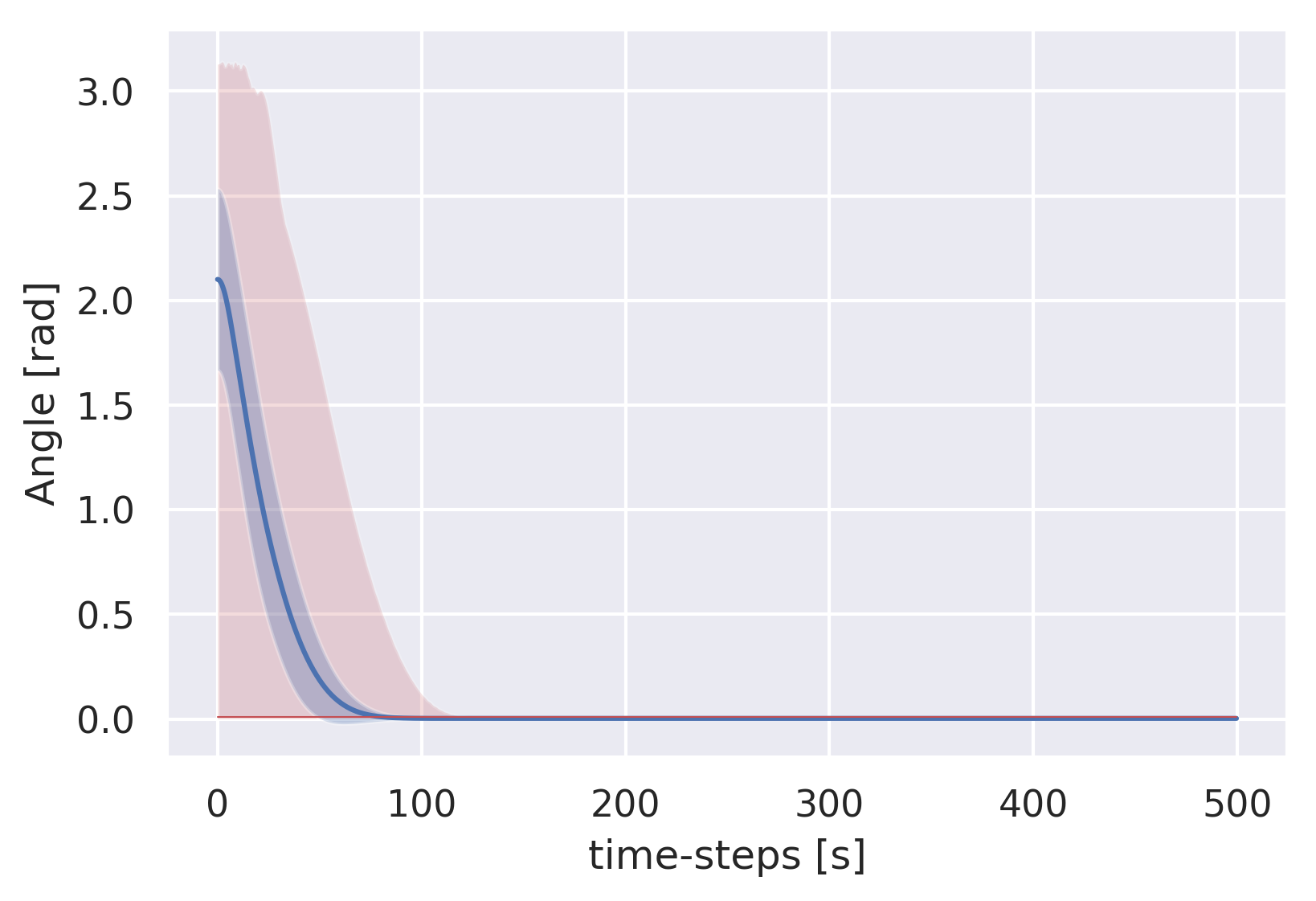}}
\caption{Agents control history for the off-nominal conditions,  tabulated in \ref{tab-failed},  with average and variance performances in blue line and blue area, and absolute worst case performances highlighted by the red area. The statistics are obtained over 10000 runs for each of the nine controllers.}
\label{failed-plots}
\end{figure}

\paragraph{Simulation results}
To assess the agents performance we chose challenging maneuvers with starting angles randomly drawn from the range $[144, 180]^\circ$ about any rotation axis from the initial orientation. The observed episode is of $1600$ steps corresponding to $800$ s, which is clipped for visualization clarity in the best performing agents to $500$ s. The accuracy threshold depends on the failed axis and the alignment vector direction. In facts, as can be noted from both Table \ref{tab-failed} and Figure \ref{failed-plots} when trying to align the satellite with the same axis where the failure occurred the performance are much higher. Note that performance varies across underactuated axes due to the satellite’s moments of inertia.
In Figure \ref{nominal-plot} the nominal controller's behaviour highlights a smooth and fast convergent control of the satellite's attitude.

\begin{figure}
\centering
\includegraphics[width=9cm, height=6cm]{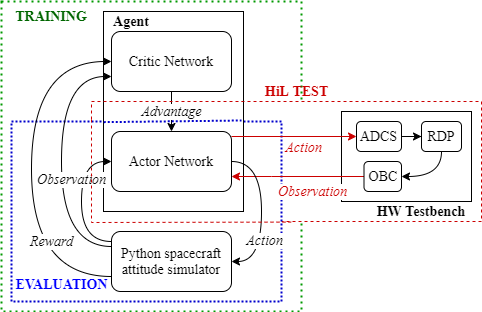}
\caption{Here the three phases of the system implementation are presented: the training of the agents which sees the interaction of the agent with the python satellite's attitude simulator using the actor-critic method, then the evaluation performed using the simulated environment and the actor network used in a deterministic fashion, finally the Hardware-in-the-Loop testing where the simulator is replaced by the ground model components of the satellite (the OBC and ADCS) along with the Real Dynamic Processor (RDP) used to propagate the attitude states.}
\label{fig:system}
\end{figure}

\paragraph{Hardware-in-the-Loop results}
The equipment used to realistically test the final control policies consists of the ArgoMoon On Board Computer (OBC), a Real Dynamic Processor (RDP) which simulates the satellite's dynamics in space with a very high degree of accuracy, the Attitude Determination and Control Subsystem (ADCS), that implements the attitude commands and generates the telemetries. To start an experiment first a list of values indicating the starting and target position, the episode time-limit, the exit conditions, and the working mode (i.e. nominal or failure on one of the three axes) is needed. As shown in Figure \ref{fig:system}, the policy networks, once the experiment is started, receive the input state from the ADCS and provide the output actions, that are used by the RDP to propagate the system's attitude, allowing the agent  to receive the new position state, closing a full real-time step in a continuous loop, which is either stopped by reaching the target precision or by meeting the exit conditions.

\begin{figure}
\centering
\subfigure[]{\includegraphics[width=6cm, height=8.5cm]{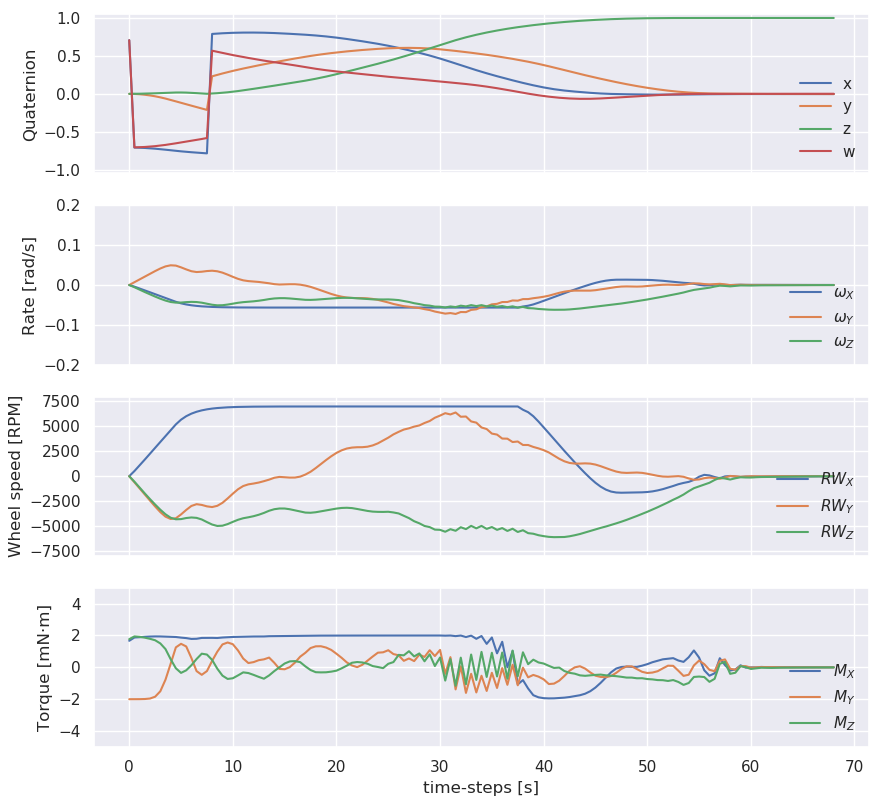}}
\subfigure[]{\includegraphics[width=6cm, height=8.5cm]{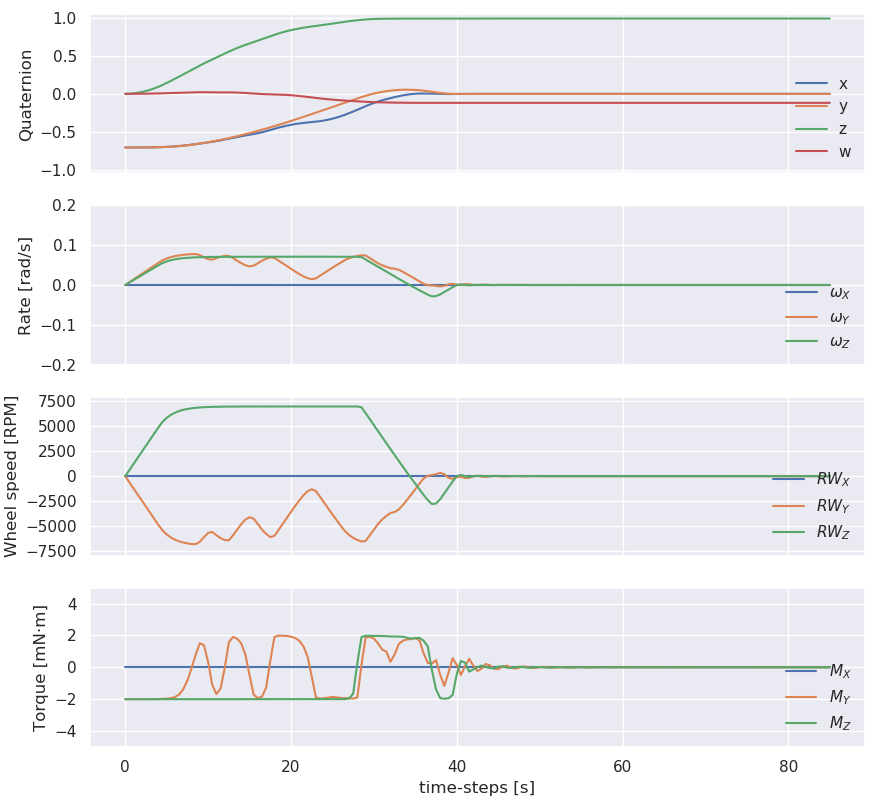}}

\subfigure[]{\includegraphics[width=6cm, height=5cm]{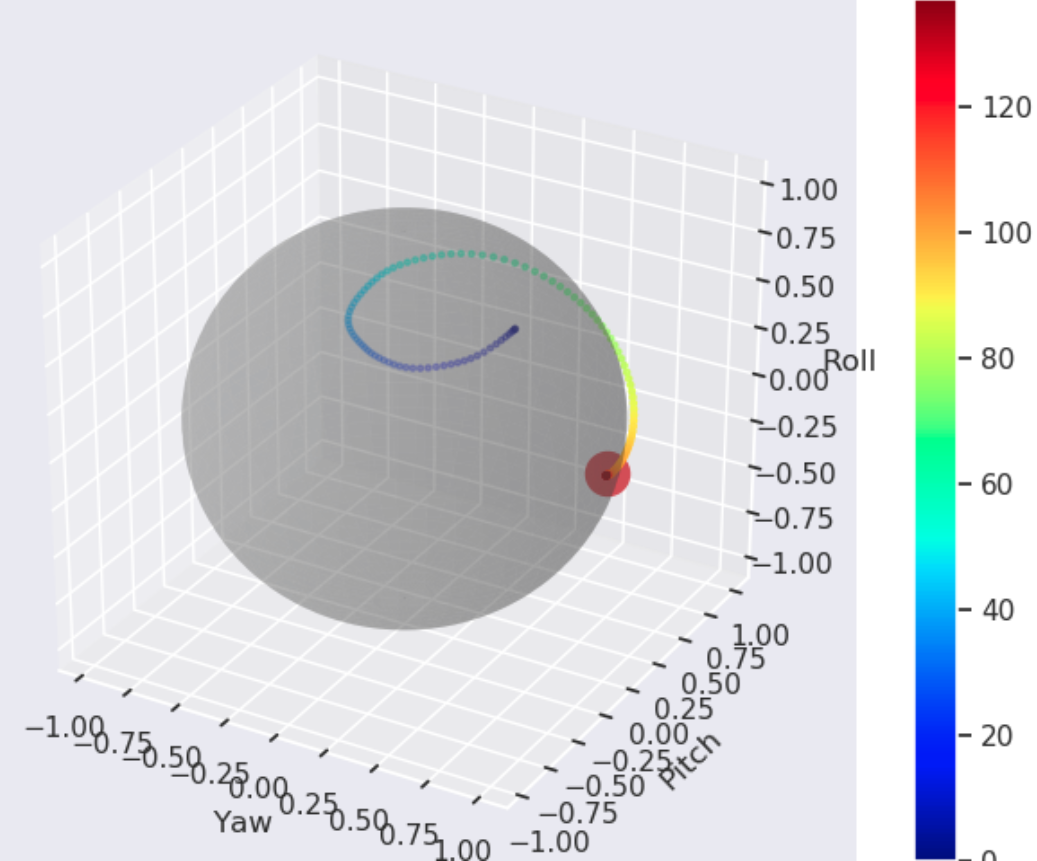}}
\hspace{0.75cm}
\subfigure[]{\includegraphics[width=6cm, height=5cm]{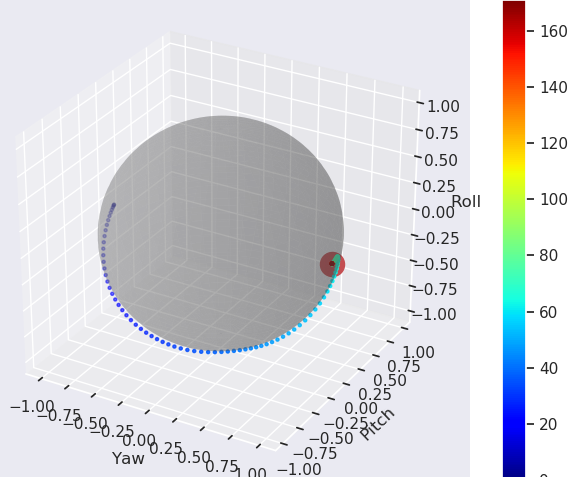}}

\vskip1ex

\caption{Agent control history for two attitude maneuvers performed using the HiL setup. The quaternion, the body rates, the RWs speeds and the commanded torques are shown in (a) and (b) for a nominal and an underactuated (with failure on the X axis) maneuver. In (c) and (d) their respective trajectories are plotted over the spherical unit quaternion.}
\label{hw-tests}
\end{figure}

The testing activity, carried out by running tens of experiments on the HiL testbench using all the different control policies, showed that the offline trained neural networks, thanks to the  injection of random delays in the training environment, were able generalize well and manage to lead the attitude trajectory towards the target regions. As shown in Figure~\ref{hw-tests} a 3D visualization of two trajectories is offered, showing the rotations performed on the sperichal representation of the quaternion. At the top part of the same figure the relevant variables are plotted for each of the trajectories, demonstrating how the sequence of torque commands leads the starting quaternion towards the target, signaled as a small red sphere in the 3D visualization. Of significant importance are the body rate and reaction wheels speed variables, which help to analyse the control behavior and to identify particularly favorable or problematic conditions. 

\begin{table}
  \caption{Agents statistics for 10000 simulated episodes.}
  \label{tab-failed}
  \centering
  \begin{tabular}{lllll}
    \toprule
    Failure & Align axis  & Accuracy [rad] & Conv. time [s] & Horizon [s]\\
    \midrule
    x   & x       & 0.01   & 72.95   & 500\\
        & y       & 0.05   & 1367.84 & 800   \\
        & z       & 0.05   & 987.06  & 800 \\
    \midrule
    y   & x       & 0.05   & 1218.91 & 800  \\
        & y       & 0.01   & 75.37   & 500 \\
        & z       & 0.05   & 1185.04 & 800  \\
    \midrule
    z   & x       & 0.05  &  1463.03  & 800 \\
        & y       & 0.05  &  1510.54  & 800  \\
        & z       & 0.01  &  142.71   & 500\\
    \bottomrule
    \end{tabular}
\end{table}

\section{Conclusion and Way Forward}\label{sec5}
We used reinforcement learning to create a flight attitude controller for a three wheel actuated small satellite. We dealt with the unavailability of a real physical simulator by using an environment with injected random noise to validate simulations and test the controller robustness. The obtained control policies were tested on representative flight hardware, evaluating with a real-time dynamic processor the maneuvers performed under realistic physical conditions 'directly on-board the satellite's HiL test-bench. 

The experiment shows that the agents could reach the target without prior knowledge of satellite parameters or working condition of the actuators, implicitly deriving the physical system properties underneath their interaction with the environment. This suggests the potential applicability, upon retraining of the parameters, of the same control strategy on platforms with different inertial mass values. Moreover, great adaptive abilities were shown by the experiments on the target hardware, highlighting that RL agents trained in simulation can transfer good behavior on realistic scenarios. 

However, RL abilities to converge are based on probabilistic assumptions which do not guarantee global stability of the control algorithms, hence the high skepticism towards applying RL to real-world problems. The authors believe that implementing an incremental approach aimed at testing in a flight scenario the control policies presented in this work, could help to learn more on the potential and the limitation of applied RL.

\section*{Acknowledgments}

The authors acknowledge the contributions of Lorenzo Bisi and Marcello Restelli in the initial phase of this study.


\begin{thebibliography}{18}
\providecommand{\natexlab}[1]{#1}
\providecommand{\url}[1]{\texttt{#1}}
\expandafter\ifx\csname urlstyle\endcsname\relax
  \providecommand{\doi}[1]{doi: #1}\else
  \providecommand{\doi}{doi: \begingroup \urlstyle{rm}\Url}\fi

\bibitem[Henna et~al.(2020)Henna, Toubakh, Kafi, and Mouchaweh]{henna2020fault}
Hicham Henna, Houari Toubakh, Mohamed~Redouane Kafi, and M.~Mouchaweh.
\newblock Towards fault-tolerant strategy in satellite attitude control
  systems: A review.
\newblock In \emph{Annual Conference of the PHM Society}, volume~12, 2020.
\newblock \doi{10.36001/phmconf.2020.v12i1.1272}.

\bibitem[Sutton and Barto(2018)]{sutton2018reinforcement}
Richard~S. Sutton and Andrew~G. Barto.
\newblock \emph{Reinforcement Learning: An Introduction}.
\newblock MIT Press, 2nd edition, 2018.

\bibitem[LeCun et~al.(2015)LeCun, Bengio, and Hinton]{lecun2015dl}
Yann LeCun, Yoshua Bengio, and Geoffrey Hinton.
\newblock Deep learning.
\newblock \emph{Nature}, 521\penalty0 (7553):\penalty0 436--444, May 2015.
\newblock \doi{10.1038/nature14539}.

\bibitem[Heess et~al.(2017)Heess, Duerig, Srinivasan, Lemmon, Merel, Wayne, and
  Silver]{heess2017elbre}
Nicolas Heess, Tobias Duerig, Sriram Srinivasan, Jay Lemmon, Josh Merel, Greg
  Wayne, and David Silver.
\newblock Emergence of locomotion behaviours in rich environments.
\newblock \emph{arXiv preprint arXiv:1707.02286}, 2017.

\bibitem[OpenAI(2018)]{openai2018ldm}
OpenAI.
\newblock Learning dexterous in-hand manipulation.
\newblock \emph{arXiv preprint arXiv:1808.00177}, 2018.

\bibitem[OpenAI et~al.(2019)OpenAI, Akkaya, Andrychowicz, Chociej, Litwin,
  McGrew, Petron, Paino, Plappert, Powell, Ribas, Schneider, Tezak, Tworek,
  Welinder, Weng, Yuan, Zaremba, and Zhang]{rubik-cube21}
OpenAI, Ilge Akkaya, Marcin Andrychowicz, Maciek Chociej, Mateusz Litwin, Bob
  McGrew, Arthur Petron, Alex Paino, Matthias Plappert, Glenn Powell, Raphael
  Ribas, Jonas Schneider, Nikolas Tezak, Jerry Tworek, Peter Welinder, Lilian
  Weng, Qiming Yuan, Wojciech Zaremba, and Lei Zhang.
\newblock Solving rubik's cube with a robot hand.
\newblock \emph{CoRR}, abs/1910.07113, 2019.
\newblock URL \url{http://arxiv.org/abs/1910.07113}.

\bibitem[Di~Tana et~al.(2019)Di~Tana, Cotugno, Simonetti, Mascetti, Scorzafava,
  and Pirrotta]{ditana2018agm}
Valerio Di~Tana, Biagio Cotugno, Simone Simonetti, Gabriele Mascetti, Edmondo
  Scorzafava, and Simone Pirrotta.
\newblock Argomoon: There is a nano-eyewitness on the sls.
\newblock \emph{IEEE Aerospace and Electronic Systems Magazine}, 34\penalty0
  (4):\penalty0 30--36, Apr 2019.
\newblock \doi{10.1109/MAES.2019.2911138}.

\bibitem[Tsiotras and Doumtchenko(2000)]{tsiotras2000csaf}
Panagiotis Tsiotras and Vassilios Doumtchenko.
\newblock Control of spacecraft subject to actuator failures: state-of-the-art
  and open problems.
\newblock \emph{Journal of Guidance, Control, and Dynamics}, 23\penalty0
  (5):\penalty0 792--802, 2000.

\bibitem[Gui et~al.(2013)Gui, Jin, and Xu]{gui2013amc}
Hong Gui, Lili Jin, and Shijie Xu.
\newblock Attitude maneuver control of a two-wheeled spacecraft with bounded
  wheel speeds.
\newblock \emph{Acta Astronautica}, 88:\penalty0 98--107, Jul 2013.
\newblock \doi{10.1016/j.actaastro.2012.03.027}.

\bibitem[Chaurais et~al.(2015)Chaurais, Ferreira, Ishihara, and
  Borges]{chaurais2015acus}
J.R. Chaurais, H.C. Ferreira, J.Y. Ishihara, and R.A. Borges.
\newblock Attitude control of an underactuated satellite using two reaction
  wheels.
\newblock \emph{Journal of Guidance, Control, and Dynamics}, 38\penalty0
  (10):\penalty0 2010--2018, 2015.

\bibitem[Zavoli et~al.(2017)Zavoli, De~Matteis, Giulietti, and
  Avanzini]{zavoli2017sapus}
A.~Zavoli, G.~De~Matteis, F.~Giulietti, and G.~Avanzini.
\newblock Single-axis pointing of an underactuated spacecraft equipped with two
  reaction wheels.
\newblock \emph{Journal of Guidance, Control, and Dynamics}, 40\penalty0
  (6):\penalty0 1465--1471, Jun 2017.

\bibitem[Tsiotras and Longuski(1995)]{tsiotras1995acass}
Panagiotis Tsiotras and James Longuski.
\newblock A novel approach to the attitude control of axisymmetric spacecraft.
\newblock \emph{Automatica}, 31:\penalty0 1099--1113, 1995.
\newblock \doi{10.1016/0005-1098(95)00010-T}.

\bibitem[Elkinsa et~al.(2020)Elkinsa, Soodb, and Rumpfc]{elkinsa2020autonomous}
Jacob~G. Elkinsa, Rohan Soodb, and Clemens Rumpfc.
\newblock Autonomous spacecraft attitude control using deep reinforcement
  learning.
\newblock \emph{Journal of Aerospace Information Systems}, 2020.
\newblock \doi{10.2514/1.I010785}.

\bibitem[Vedant et~al.(2019)Vedant, Allison, West, and Ghosh]{vedant2019rl_ac}
M.~Vedant, James~T. Allison, Matthew West, and Alexander Ghosh.
\newblock Reinforcement learning for spacecraft attitude control.
\newblock \emph{Proceedings of the International Astronautical Congress (IAC)},
  2019.

\bibitem[Xu et~al.(2019)Xu, Wu, and Zhao]{xu_model-based_2019}
Ke~Xu, Fengge Wu, and Junsuo Zhao.
\newblock Model-based deep reinforcement learning with heuristic search for
  satellite attitude control.
\newblock \emph{Industrial Robot: An International Journal}, 46\penalty0
  (3):\penalty0 415--420, 2019.
\newblock \doi{10.1108/IR-05-2018-0086}.

\bibitem[Brockman et~al.(2016)Brockman, Cheung, Pettersson, Schneider,
  Schulman, Tang, and Zaremba]{openAI_gym}
Greg Brockman, Vicki Cheung, Ludwig Pettersson, Jonas Schneider, John Schulman,
  Jie Tang, and Wojciech Zaremba.
\newblock Openai gym.
\newblock \emph{CoRR}, abs/1606.01540, 2016.
\newblock URL \url{http://arxiv.org/abs/1606.01540}.

\bibitem[Paszke et~al.(2019)Paszke, Gross, Massa, and et~al.]{pytorch2019}
Adam Paszke, Sam Gross, Francisco Massa, and et~al.
\newblock Pytorch: An imperative style, high-performance deep learning library.
\newblock In H.~Wallach, H.~Larochelle, A.~Beygelzimer, F.~d'Alché Buc,
  E.~Fox, and R.~Garnett, editors, \emph{Advances in Neural Information
  Processing Systems 32}, pages 8024--8035. Curran Associates, Inc., 2019.
\newblock URL \url{http://papers.neurips.cc/paper/9015}.

\bibitem[Schulman et~al.(2017)Schulman, Wolski, Dhariwal, Radford, and
  Klimov]{schulman2017ppo}
John Schulman, Filip Wolski, Prafulla Dhariwal, Alec Radford, and Oleg Klimov.
\newblock Proximal policy optimization algorithms.
\newblock \emph{arXiv preprint arXiv:1707.06347}, 2017.

\end{thebibliography}

\appendix
\section{Appendix}
\subsection{Quaternions for attitude representation}\label{appendix-quaternion}
Even though Euler angles are easier to visualize, quaternions are preferred because they do not suffer from the Gymbal Lock problem and they are less computational intensive. A quaternion is a $4 \times 1$ matrix with a scalar element and a vector part, which, according to Euler's rotational theorem, expresses a rotational angle around a normalized vector as follows:
\begin{equation}
    q =
        \begin{bmatrix}
    q_s \\
    q_x \\
    q_y \\
    q_z \\
    \end{bmatrix}
    =
     \begin{bmatrix}
    s \\
    v_x \\
    v_y \\
    v_z \\
    \end{bmatrix}
    =
     \begin{bmatrix}
    s \\
    \overset{\rightarrow}{v} 
    \end{bmatrix}
    =
    \begin{bmatrix}
    \cos(\frac{\theta}{2}) \\
    \norm{\overset{\rightarrow}{v}} * \sin(\frac{\theta}{2})
    \end{bmatrix}
    \label{eq:quaternion}
\end{equation}

\subsection{Attitude propagation}\label{appendix-propagation}

The propagation of the attitude is performed assuming that quaternion at time t + $\Delta$t is given by composition of quaternion at time t and a quaternion representing the rotation in the interval $\Delta$t, as shown in equation \ref{eq:quat_rate}. Jointly integrating the equations \ref{eq:euler} and \ref{eq:quat_rate} using the Euler method with a step size of $100$ we can both obtain the quaternion and body rates for each time-step.

\begin{equation}
    \Omega =
        \begin{bmatrix}
    0         &   \omega_x  &  -\omega_y  &  \omega_z \\
    -\omega_x &   0         & \omega_z    &  \omega_y  \\
    \omega_y  &   -\omega_z &   0         &  \omega_x \\
    -\omega_z &   -\omega_y &   -\omega_x &   0        \\
    \end{bmatrix}
    \label{eq:omega}
\end{equation}

\begin{equation}
 q(t+\Delta t) = \left[ I+\tfrac{1}{2}~\Omega~\Delta t\right] q(t)
    \label{eq:quat_rate}
\end{equation}

\begin{equation}
 \Delta q = \tfrac{1}{2}~\Omega~q(t)
    \label{eq:quat_derivation}
\end{equation}

In equation \ref{eq:quat_derivation} $\Delta q$ corresponds to the quaternion error, i.e. the rotation needed to get from the actual to the target position, while $\Omega$, as expressed in equation \ref{eq:omega}, is the rotational speed reference with respect to the inertial frame.

\subsection{Neural Networks hyperparameters}\label{appendix-hyperparam}

The hyperparameters used for the models training can be found in the following table.

\begin{table}
  \caption{PPO hyperparameters used for training}
  \label{tab-ppo_hyperparameters}
  \centering
  \begin{tabular}{ll}
    \toprule
    Hyperparameter &    Value\\
    \midrule
     Discount ($\gamma$)  & 0.99  \\
     KL target            & 0.035   \\
     Epochs $\#$            & 40   \\
     Learning rate ($\alpha$)  & 0.0003  \\
     Batch size         & 150  \\
     Minibatch size     & 32   \\
     Clipping parameter ($\epsilon$) & 0.2  \\
    \bottomrule
    \end{tabular}
\end{table}

\end{document}